\documentclass[runningheads]{llncs}

\usepackage[T1]{fontenc}
\usepackage{graphicx}
\usepackage{multirow}%
\usepackage[title]{appendix}%
\usepackage{xcolor}%
\usepackage{textcomp}%
\usepackage{manyfoot}%
\usepackage{booktabs}%
\usepackage{algorithm}%
\usepackage{algorithmicx}%
\usepackage{algpseudocode}%
\usepackage{listings}%
\usepackage{url}
\usepackage{hyperref}
\usepackage[utf8]{inputenc}

\usepackage{biblatex} 
\usepackage[acronym]{glossaries}
\usepackage{enumitem,kantlipsum}

\begin{document}
\title{Vehicle Detection Performance in Nordic Region }
%
%
\author{Hamam Mokayed\inst{1} \and
Rajkumar Saini\inst{1}\and
Oluwatosin Adewumi\inst{1}\and
Lama Alkhaled\inst{1}\and
Björn Backe\inst{1}\and
Palaiahnakote Shivakumara\inst{2}\and
Olle Hagner\inst{3} \and
Yan Chai Hum\inst{4}}
\authorrunning{H. Mokayed et al.}
 \institute{Electrical and Space Engineering, Luleå University of Technology, Luleå, Sweden, \email{fname.last@ltu.se} 
 \and
 School of Science, Engineering and Environment, Salford university, UK
\email{S.Palaiahnakote@salford.ac.uk}
\and
 Smartplanes, Jävre, Sweden, 
 \email{olle.hagner@smartplanes.se} 
 \and
 Mechatronics and Biomedical Engineering, Universiti Tunku Abdul Rahman, Selangor, Malaysia
 \email{humyc@utar.edu.my}}
\maketitle              
\begin{abstract}
This paper addresses the critical challenge of vehicle detection in the harsh winter conditions in the Nordic regions, characterized by heavy snowfall, reduced visibility, and low lighting. Due to their susceptibility to environmental distortions and occlusions, traditional vehicle detection methods have struggled in these adverse conditions. The advanced proposed deep learning architectures brought promise, yet the unique difficulties of detecting vehicles in Nordic winters remain inadequately addressed. This study uses the Nordic Vehicle Dataset (NVD), which has UAV images from northern Sweden, to evaluate the performance of state-of-the-art vehicle detection algorithms under challenging weather conditions. Our methodology includes a comprehensive evaluation of single-stage, two-stage, and transformer-based detectors against the NVD. We propose a series of enhancements tailored to each detection framework, including data augmentation, hyperparameter tuning, transfer learning, and novel strategies designed explicitly for the DETR model. Our findings not only highlight the limitations of current detection systems in the Nordic environment but also offer promising directions for enhancing these algorithms for improved robustness and accuracy in vehicle detection amidst the complexities of winter landscapes. The code and the dataset are available at \url{https://nvd.ltu-ai.dev}.

\end{abstract}
\section{Introduction}
Vehicle detection systems are crucial for many applications, including traffic management and autonomous navigation. Yet, their performance in adverse weather conditions, especially in the harsh winter of the Nordic regions, presents significant challenges. The consistent snowfall, reduced visibility, and low lighting conditions inherent to these areas complicate vehicle detection tasks, demanding an evaluation and enhancement of current detection methods to ensure reliability and robustness \cite{sakhare2020review}. Initially, traditional non-deep learning techniques were employed for this task. However, these conventional methods faced challenges due to susceptibility to image distortions, vehicle occlusions, and variations in illumination. As a result, their accuracy remained limited, constraining their applicability to specific scenarios  \cite{mokayed2021new, mokayed2023nordic, tsai2007vehicle,felzenszwalb2008discriminatively, howard2017mobilenets, mokayed2022license}. With the advent of deep learning, various approaches aimed to enhance vehicle detection by introducing more advanced network architectures \cite{geiger2012we,howard2017mobilenets,hu2021utilising}. The broader field of object detection has witnessed extensive research, yet many of these general-purpose classifiers struggle to achieve competitive performance in vehicle detection benchmarks. This is due to the unique challenges posed by vehicle detection, including significant light variations, dense occlusions, and size disparities. These challenges prompted us to explore the performance of various vehicle detection architectures under such demanding conditions. The Nordic Vehicle Dataset (NVD) \cite{mokayed2023nordic} serves to address and highlight these challenges. Capturing 22 aerial videos across northern Sweden's snowy terrain, the NVD provides a comprehensive view of the difficulties faced in vehicle detection from unmanned aerial vehicles (UAVs). 

\subsection{UAV Dataset}
A search on the various available UAV datasets is conducted to validate the originality and value our study brings to this domain by evaluating and trying to improve the performance of different vehicle detectors using the NVD dataset. 
Several UAV datasets have been compiled to tackle orientation and scale issues, primarily focusing on capturing vehicles on roads in clear weather conditions. \cite{mustafa2024unmanned, wang2019orientation, bemposta2022dataset}.
The VisDrone dataset\cite{cao2021visdrone} focuses broadly on object detection using drone imagery but lacks specificity in weather conditions and vehicle detection. UAV project dataset \cite{wan2021vistrongerdet} comes closer to our approach but is limited to foggy conditions and road-bound vehicles. The Mimos drone dataset \cite{mokayed2021new,mokayed2014car} is collected in clear weather with a different focus on detecting the plate numbers. The Flickr and CARISSMA datasets \cite{rothmeier2021let} are outcomes of research to create a model capable of applying simulated weather effects to images, a concept that diverges from our research objectives. These datasets comprise images covering various atmospheric conditions, such as fog, rain, and snow, including vehicle imagery. However, they primarily focus on street-level footage and lack UAV-captured perspectives, leading to a significant difference in viewing angles. The Video-Traffic dataset \cite{liu2015fast} wants to create a model to gather traffic information, which means it only looks at vehicles on the streets. It doesn't say anything about the weather we want to investigate. This dataset was taken by drones flying over Chongqing's main roads at a height of 200-250 meters. The videos are very clear, with a resolution of 3840 by 2160 pixels, and were made following the VOC2007 standard. Data synthesizing research offers relevant insights but differs significantly in focus and methodology, with the former concentrating on text detection on vehicles and the latter on generating adverse weather conditions on images. Lastly, the UAV videos for traffic study, though aligned with our general goal of vehicle detection via UAV, are narrowly focused on traffic information on city roads without considering weather impacts. Overall, our research stands out by aiming for comprehensive vehicle detection in any location, under challenging weather conditions, leveraging a dataset annotated with detailed precision.

\subsection{Vehicle Detectors}
Numerous techniques have been devised previously to tackle the complexities associated with vehicle detection, particularly concerning small-sized vehicles and scenarios involving multiple vehicles within images in different weather conditions. Vehicle detectors have evolved with significant advancements in accuracy, speed, and robustness. The most prominent types of vehicle detectors include single-stage detectors, two-stage detectors, and transformer-based detectors, each with distinct architectures and operational mechanisms \cite{zhao2023revisiting}.\newline
\begin{enumerate}[wide, labelindent=0pt]
\item Single stage (or single pass): these model architectures use the neural network in predicting the bounding boxes and class probabilities of objects in one step for a full image input \cite{girshick2014rich,lohia2021bibliometric}.
Examples of such models include the popular You Only Look Once (YOLO) \cite{redmon2016you}, Fast Detection \cite{hossain2020fast}, and 3D-DETNet \cite{li20183d}.
They tend to have high inference speeds \cite{lohia2021bibliometric}.
These models have been applied in vehicle detection tasks \cite{wang2020soft,li20183d}.
The introduction of the SotA YOLO, now in version 9 \cite{wang2024yolov9}, marked a milestone in computer vision.
The latest version combines two novel concepts: Programmable Gradient Information (PGI) and Generalized Efficient Layer Aggregation Network (GELAN).
\par\hspace{\dimexpr-\leftmargin-\listparindent}

\item Double stage (or double pass): these model architectures generate candidate regions before a second stage of pooling operation to classify these regions \cite{lohia2021bibliometric,meng2020vehicle}.
Examples of these models are Region-based Convolutional Neural Network (R-CNN) \cite{girshick2014rich} and Faster R-CNN \cite{ren2015faster}, and FPN \cite{lin2017feature}.
They tend to have high recognition accuracy and are also used for vehicle detection \cite{lohia2021bibliometric,yang2019vehicle}.
The Faster R-CNN model introduced a Region Proposal Network (RPN), which shares convolutional features of the full input image with the detection network. 
\par\hspace{\dimexpr-\leftmargin-\listparindent}

\item Transformers-based detectors: these detection models use self-attention mechanisms, allowing each part of the image to be considered in the context of every other part. This enhances the model's ability to understand complex scenes and the relationships between different objects within them. This global processing capability is particularly beneficial in densely populated scenes or scenarios where objects' context and relative positioning are crucial for accurate detection. Transformer generalizes well to other tasks by applying it successfully to English constituency parsing both with large and limited training data. The Transformer can be trained significantly faster than architectures based on recurrent or convolutional layers \cite{vaswani2017attention}. Examples of these models, Detection Transformer (DETR) \cite{carion2020end}, Detection Transformer-Spatial Pyramid Pooling (DETR-SPP) \cite{sp2023detr},demformable DETR \cite{zhu2020deformable}, and Swin Transformer \cite{liu2021swin}.
\end{enumerate} 

\section{Proposed Method}
\begin{figure}[H]
\centering
\includegraphics[width=0.8\textwidth]{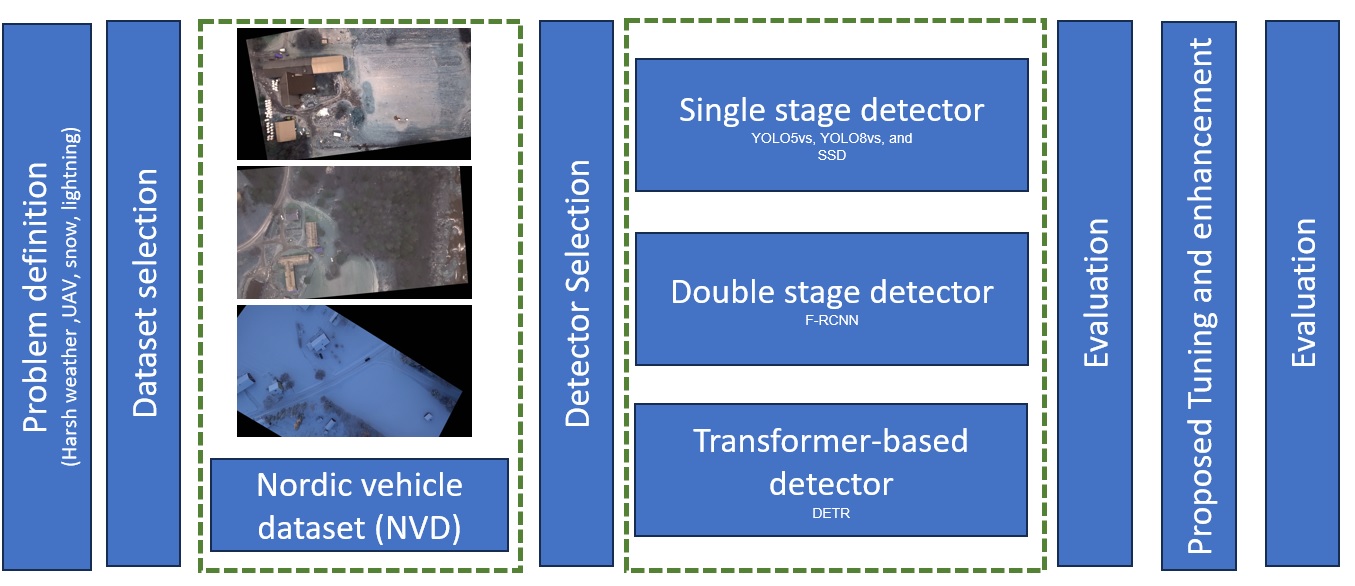}
\caption{Proposed Method.} \label{Proposed method}
\end{figure}

The foundation of the proposed method for evaluating vehicle detection performance in Nordic environments hinges on the strategic selection of the Nordic Vehicle Dataset (NVD) as a rigorous testing ground. With its comprehensive compilation of challenging scenarios collected from the Nordic region, the NVD provides a unique opportunity to validate, tune, and enhance various state-of-the-art (SOTA) vehicle detection algorithms. This methodological approach is designed to encompass a broad spectrum of contemporary detection frameworks, categorically spanning single-stage, two-stage, and transformer-based detectors, each known for their distinct operational paradigms and performance characteristics, refer to Fig~\ref{Proposed method}.

\subsection{Dataset Selection: Nordic Vehicle Dataset (NVD)}
The NVD's rich repository of UAV images, meticulously captured over the snowy landscapes of northern Sweden, provides a comprehensive view of the difficulties faced in vehicle detection from unmanned aerial vehicles (UAVs) in harsh weather. With altitudes ranging from 120 to 250 meters, the dataset encompasses a variety of snow and cloud conditions across 8,450 annotated frames featuring 26,313 cars. The diverse video resolutions, frame rates, and varying Ground Sample Distance (GSD) metrics offer a detailed representation of vehicles against the challenging backdrop of Nordic winters \cite{mokayed2023nordic}. 

\subsection{Evaluation Framework}
The method entails a structured evaluation framework wherein a diverse array of SOTA vehicle detection algorithms will be systematically tested against the NVD.

\begin{enumerate}[wide, labelindent=0pt]
\item Single-Stage Detectors: 
YOLO5vs, YOLO8vs, and SSD (Single Shot MultiBox Detector) will be assessed for their detection capabilities in the face of the NVD's challenging conditions.
\par\hspace{\dimexpr-\leftmargin-\listparindent}
\item Two-Stage Detectors: 
Faster R-CNN (F-RCNN) will undergo rigorous testing to determine its effectiveness in accurately identifying vehicles amidst heavy occlusions and varying snow cover.
\par\hspace{\dimexpr-\leftmargin-\listparindent}
\item Transformer-Based Detectors: 
DETR  will be evaluated for their ability to model long-range dependencies and complex scene contexts.
\end{enumerate} 

\subsection{Performance Tuning and Enhancement}
Critical to this proposed method is the iterative process of tuning and enhancing the algorithms based on their performance metrics against the NVD. This involves not only the adjustment of hyperparameters but also the potential integration of novel steps specifically tailored to overcome the identified challenges by NVD.
\begin{enumerate}[wide, labelindent=0pt]
\item Single-Stage Detectors: 
Improving the performance of SSD and YOLO in general, regardless of the model version, involves a series of traditional techniques and strategies. In this work, we will focus on the following \cite{maity2021faster,chung2020data}.
\begin{enumerate}
\item Data augmentation is a critical step that involves enriching the training dataset through various operations to improve the model's generalization ability.
\item Hyperparameter tuning is essential for adjusting different parameters such as learning rates, batch sizes, and epochs.
\item Transfer learning is an effective practice used to utilize pretrained models on larger datasets.
\end{enumerate}
\par\hspace{\dimexpr-\leftmargin-\listparindent}
\item Two-Stage Detectors: 
Improving the performance of F-RCNN involves a series of traditional techniques and strategies. In this work, we will focus on the following \cite{mo2020improved}:
\begin{enumerate}
\item Data augmentation.
\item Hyperparameter tuning.
\end{enumerate} 
\par\hspace{\dimexpr-\leftmargin-\listparindent}
\item Transformer-Based Detectors: This paper primarily focuses on introducing enhancements to the Detection Transformers (DETR) model to improve its performance. In this regard, the following steps will be presented:  
\begin{enumerate}[wide, labelindent=0pt]
\item Initial Region Identification under Adverse Conditions:\newline
Initially, a robust algorithm called maximally stable extremal regions (MSER) is proposed, which generates a novel set of image components known as extremal regions. These regions are characterized by two features derived from the projection transformation of image coordinates and the monotonic transformation of image intensities \cite{matas2004robust}. Affine invariant feature descriptors are computed on a grayscale image, and although MSER's robustness varies from multiple measurement regions derived from invariant constructs from extremal regions \cite{donoser2006efficient}, certain regions exhibit distinct characteristics that are notably larger and potentially useful for establishing preliminary correspondences \cite{matas2004robust}. As MSER can generate numerous blobs of varying sizes, accommodating original image resolution detection as well as different resolutions stemming from long distances or blurred (coarse) images, it leads to a loss of image details and connections between different regions and their neighbors \cite{donoser2006efficient}. MSER's strength lies in its capability to maintain invariance to scale changes in the scene image across different resolutions, thus stabilizing vehicle regions. The image's resolution is obtained using a scale pyramid (without Gaussian filtering), encompassing one octave per scale and a total of three scales ranging from the finest image (input image) to the coarser, blurred image. This process results in the creation of multiresolution maximally stable extremal regions, denoted as MR-MSER, which are subsequently applied to NVD's images \cite{matas2004robust} as clarified in Figure ~\ref{MSER}).
\begin{figure}[H]
\centering
\includegraphics[width=\textwidth]{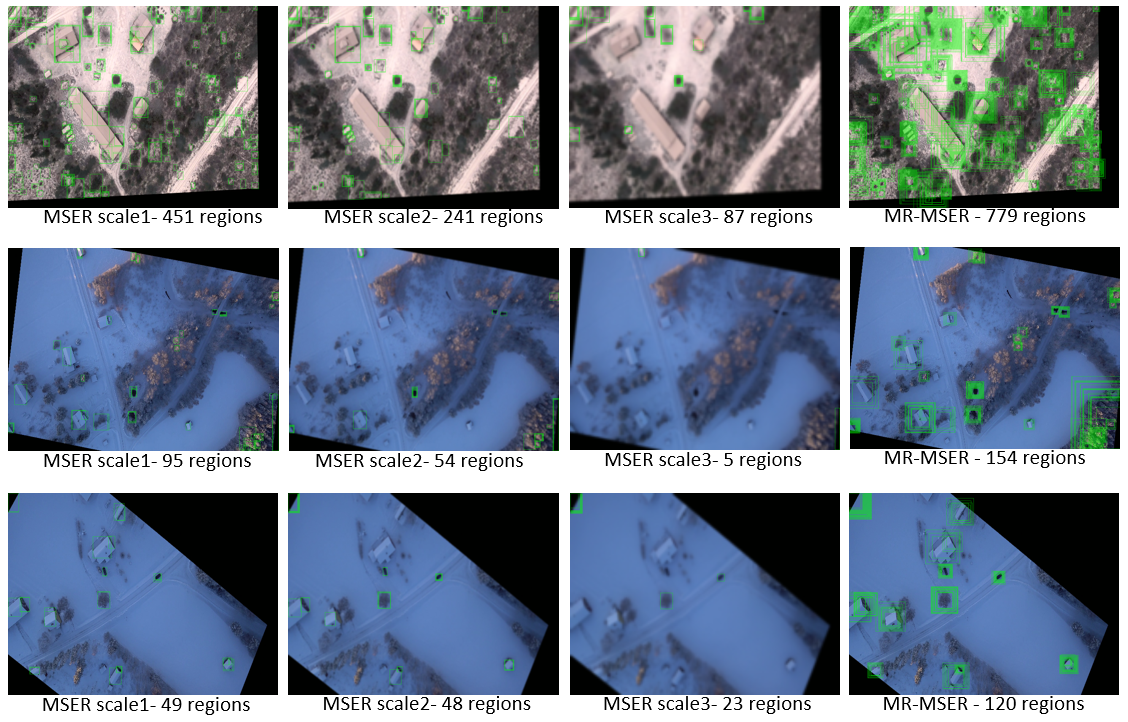}
\caption{Initial Region Identification by MSER and MR-MSER.} \label{MSER}
\end{figure}

To enhance region detection recall, an augmentation process is applied to MR-MSER. The augmentation procedure is implemented for each MR-MSER region. Each MR-MSER region of varying sizes is enclosed within a rectangular bounding box that remains centered. This bounding box is then transformed into a square shape by expanding its area by 30\% and 60\% in three distinct dimensions. These different square dimensions are resized to a 28×28 pixel image patch. Each image patch undergoes random rotation within the range of [ $\pi$/4,  $\pi$/4] four times, facilitating model training \cite{zamberletti2015augmented}. The regions generated by applying augmented MR-MSER over NVD's images are shown in Figure ~\ref{MSER}.

\item  Feature Analysis and Reduction in Obscured Environments.\newline
The purpose of using the rough set approach is knowledge discovery and approximation of sets using granular information. By applying RST, one can reduce the dimensionality of the feature space and generate decision rules that are crucial for distinguishing between vehicle and non-vehicle regions. This step enhances the computational efficiency and effectiveness of the subsequent detection process. Given a confidence map, the process for granularization splits the input image window into multiple windows with a resolution of each sub-window (g = 4). The main purpose is to classify pixel values into vehicle and nonvehicle approximations. Let a set of objects be U. There is also an indiscernibility relation R $\subseteq$ U* U that refers to the central concept of rough set theory. In the indiscernibility relation, the values of the object are identical, considering a subset of the related attributes. In other words, it is an equivalence relation where all identical values of the object are elementary. Hence, R can also be considered an equivalence relation. Let X be a subset of U with two possibilities: either is crisp, which is explicit with respect to R if the boundary region of X is empty, or is rough, which is in-explicit with respect to R if the boundary region of X is nonempty using RST to characterize the set U as possible for lower approximation and upper approximation, and boundary region of set X.
 \begin{equation}
\underline{R}(x)\ =\ \bigcup_{x\in U}{R(x):R(x)\subseteq X}   \label{eq1}
\end{equation}
R-upper approximation of X:
\begin{equation}
\overline{R}(x)=\bigcup_{x\in U}{R(x):R(x)\cap X\neq\phi}\    \label{eq2}
\end{equation}
R-boundary region of X:
\begin{equation}
\overline {RN}_R(X)=\overline{R}(X)\ -\ \underline{R}(X) 
\end{equation}

Rough entropy (RE) is introduced, which can avoid imprecision to find the optimum threshold as precisely as possible. The rough entropy threshold (RET) as the reference for the threshold in the binarization approach in the grayscale image, which was obtained by using a sliding window with a nonoverlapping granule window in m×n, is set as a 2×2 window size. RET can be defined as \cite{pal2005granular}.
\begin{equation}
{RE}_T=-\frac{\mathcal{\exp}}{2}\left[R_{OT}{log}_\mathcal{\exp}{\left(R_{OT}\right)+R_{BT}{log}_\mathcal{\exp}{\left(R_{BT}\right)}}\right]
\end{equation}
where
$R_{OT}=1-\frac{\left|{\overline{O}}_T\right|}{\left|{\underline{O}}_T\right|}$ is the roughness of the object, $R_{BT}=1-\frac{\left|{\overline{B}}_T\right|}{\left|{\underline{B}}_T\right|}$ is the roughness of the background, $\left|{\overline{O}}_T\right|$ and $\left|{\underline{O}}_T\right|$ are the cardinality of the sets ${\overline{O}}_T$ and ${\underline{O}}_T$ for a given image depending on the value T, and $\left|{\overline{B}}_T\right|$ and $\left|{\underline{B}}_T\right|$ are the cardinality of the sets ${\overline{B}}_T$ and ${\underline{B}}_T$ for a given image depending on the value T.\\
\\The principle of reducing the roughness of both the object and background and maximizing ${RE}_T$ is computed for every T representing the object and background regions, respectively (0,. . . ,T) and (T+1,. . . ,L-1). The optimum threshold is selected for the maximum ${RE}_T$ to provide the object-background segmentation given by the definition of $T^\ast$.
\begin{equation}
T^\ast=\arg\max_T{{RE}_T}
\end{equation}
Maximizing the rough entropy ${RE}_T$ to obtain the required threshold implies minimizing both the object roughness and background roughness such that this method is an object enhancement/extraction method \cite{pal2005granular}.

The confidence map is generated to show the final regions selected after applying RST as the filtration layer. This confidence map is constructed by utilizing confidence values from each stacked regions. Regions with higher intensity in the confidence map are indicative of potential vehicle components. The outcome of the generated augmented confidence map is shown in Figure~\ref{RST}.

\begin{figure}[H]
\centering
\includegraphics[width=\textwidth]{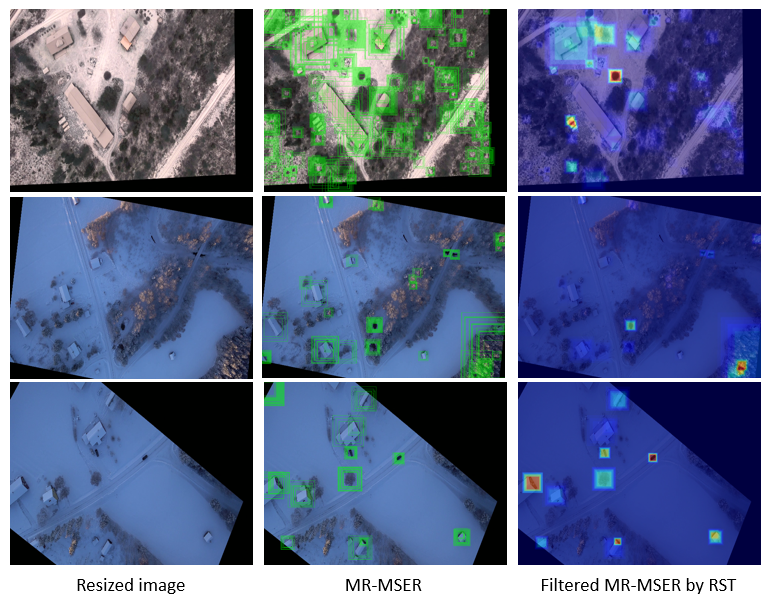}
\caption{Filtered region Identification by RST and MR-MSER.} \label{RST}
\end{figure}

\item Refined DETR Detection in Complex Contexts \newline
Maximally Stable Extremal Regions (MSER) refined by rough set, a method renowned for its robustness in detecting coherent regions in images, presents a promising solution to enhance DETR's capabilities in these complex visual environments. By integrating refined MSER with DETR, the improved model can leverage the strength of MSER in efficiently segmenting and identifying stable regions within images, even under severe weather distortions, as shown in Figure~\ref{RDETR}. This fusion aims to provide a more resilient feature extraction mechanism, allowing DETR to better recognize and localize objects partially or fully obscured by snow.

\begin{figure}[H]
\centering
\includegraphics[width=\textwidth]{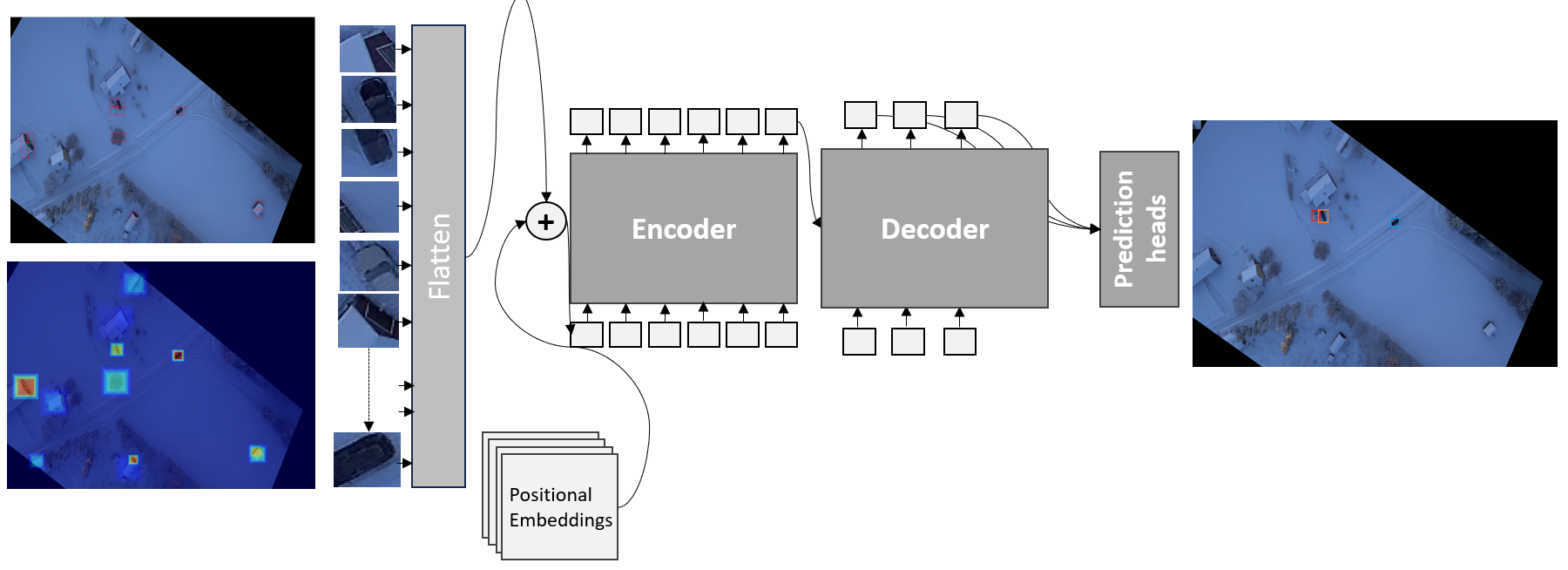}
\caption{Refined DETR performance.} \label{RDETR}
\end{figure}

\end{enumerate}
\end{enumerate}

\section{Experimental Results}

\begin{figure}[]
\centering
\includegraphics[width=0.8\textwidth]{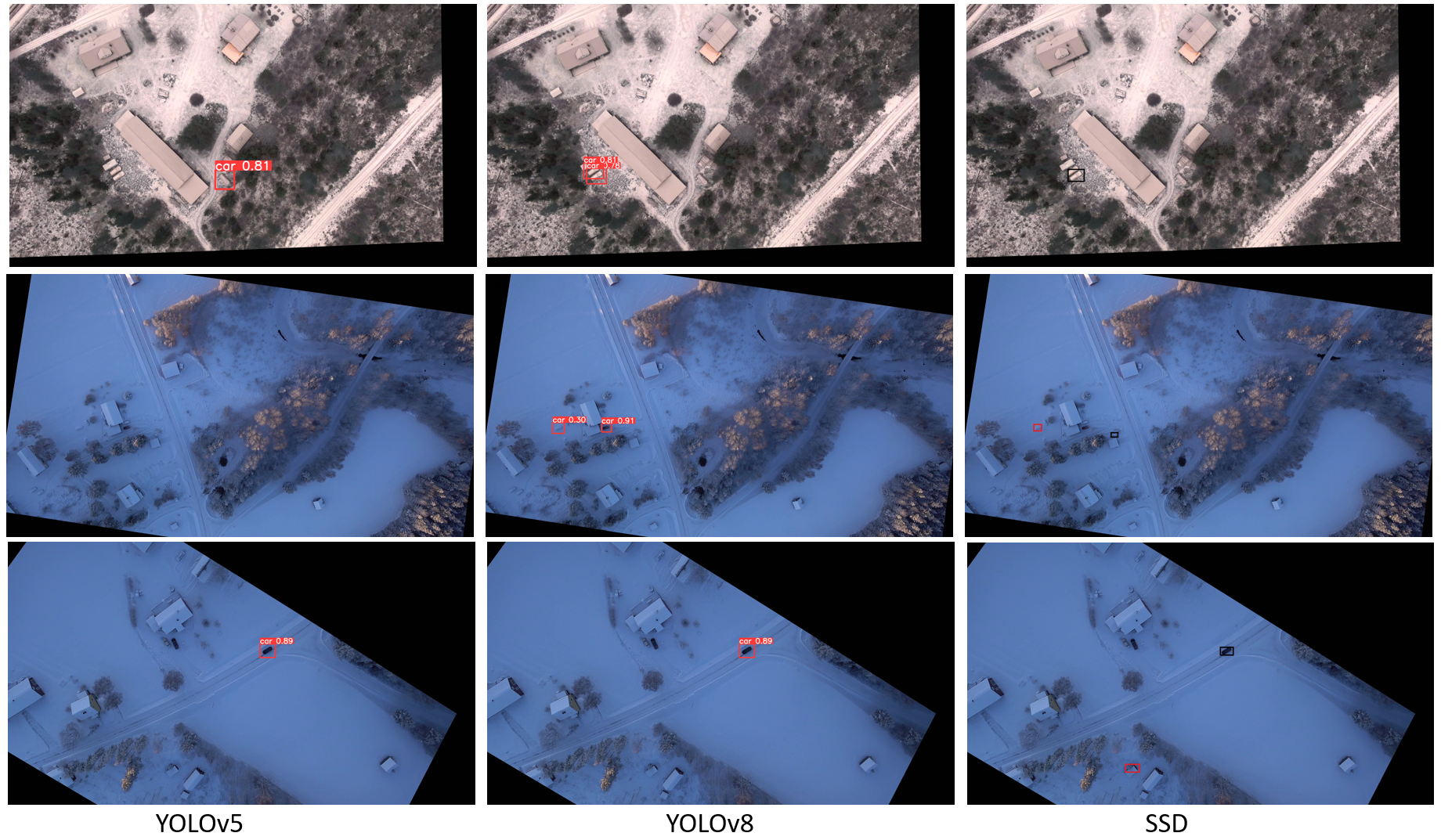}
\caption{Single detectors performance without enhancement.} \label{Single detector ini}
\end{figure}

\begin{figure}[H]
\centering
\includegraphics[width=0.8\textwidth]{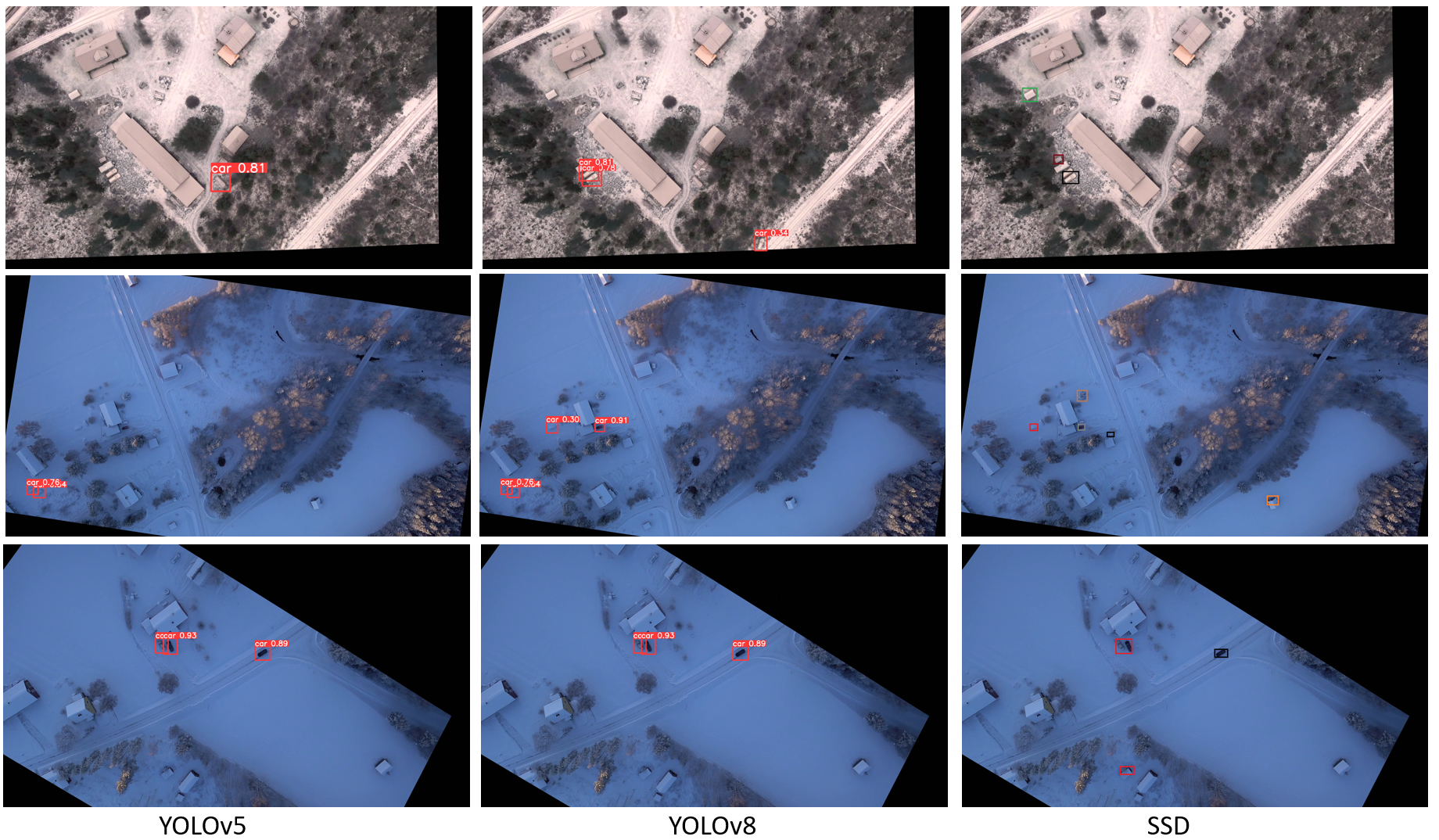}
\caption{Single detectors performance with the proposed methodology.} \label{Single detector}
\end{figure}

This section presents the results obtained from experiments conducted using different detectors before and after implementing specific performance enhancements customized for each detector. The experiments initially showcase the performance of single-stage detectors without any enhancements, as illustrated in Figure~\ref{Single detector ini}, along with the corresponding accuracy detailed in Table~\ref{tab1}. Subsequently, the enhancement techniques proposed for single detectors, as described in the methodology, are implemented to demonstrate improvements in detector performance, as depicted in Figure~\ref{Single detector}, Table~\ref{tab3}.\newline
\textbf{1. Single-Stage Detectors}
\begin{table}[h]
\centering
\caption{Model accuracy without enhancement}\label{tab1}
\begin{tabular}{|l|l|l|l|l|}
\hline
Model &  Precision & Recall & mAP50 & mAP50-95 \\
\hline
YOLOv5s	& 69.00\%	&32.10\% &	53.20\%&	31.70\% \\
YOLOv8s&	72.40\%	&28.00\%	&45.80\%	&22.80\% \\
SSD&	31.20\%	&18.00\%	&26.8\%	&12.4\% \\
\hline
\end{tabular}
\end{table}

\begin{table}[H]
\centering
\caption{Model accuracy by implementing the proposed methodology}\label{tab3}
\begin{tabular}{|l|l|l|l|l|}
\hline
Model &  Precision & Recall & mAP50 & mAP50-95 \\
\hline
YOLOv5s	& 70.6\%	&48.2\% &	56.0\%&	33.80\% \\
YOLOv8s&	77.1\%	&34.60\%	&50.7\%	&24.22\% \\
SSD&	39.60\%	&25.8\%	&33.75\%	&20.2\% \\
\hline
\end{tabular}
\end{table}

\begin{figure}[H]
\centering
\includegraphics[width=0.78\textwidth]{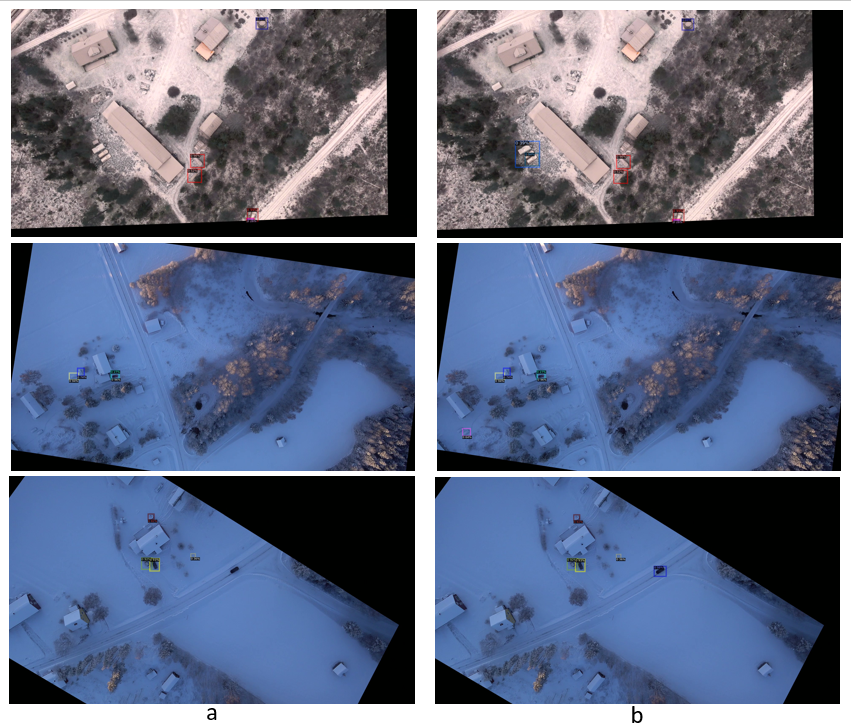 }
\caption{Double detector performance a)without enhancement, b)with enhancement} \label{Double detector ini}
\end{figure}

The study assessed the effectiveness of a dual-stage detector, exemplified by F-RCNN, both prior to and after the implementation of suggested enhancements, as illustrated in Figure~\ref{Double detector ini}, Table~\ref{tab4}.\newline
\textbf{2. Two-Stage Detectors }

\begin{table}[H]
\centering
\caption{F-RCNN performance}\label{tab4}
\begin{tabular}{|l|l|l|l|l|}
\hline
Model &  Precision & Recall & mAP50 & mAP50-95 \\
\hline
F-RCNN no enhancement	& 8.4\%	&13.3\%	&10.15\%	&4.30\% \\
\hline
F-RCNN with enhancement	& 22.40\%	&26.3\%	&23.25\%	&12.10\% \\
\hline
\end{tabular}
\end{table}


\begin{figure}[H]
\centering
\includegraphics[width=0.8\textwidth]{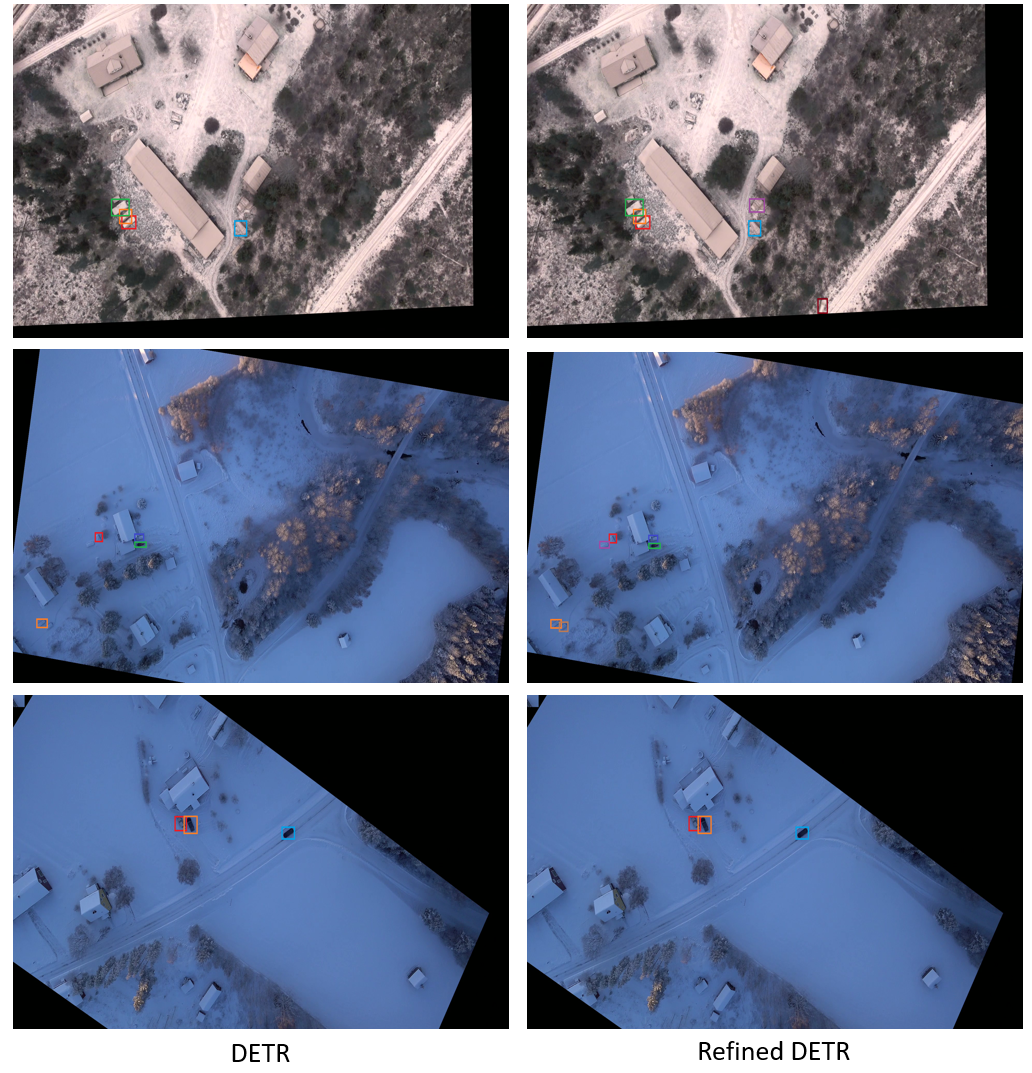 }
\caption{DETR vs Refined DETR } \label{RDETR2}
\end{figure}

Finally, we evaluated the performance of both DETR and an enhanced version of DETR incorporating MR-MSER and rough set theory, as described in Figure~\ref{RDETR2}, Table~\ref{tab5}.\newline
\textbf{3. Transformer-Based Detectors}

\begin{table}[H]
\centering
\caption{DETR Performance }\label{tab5}
\begin{tabular}{|l|l|l|l|l|}
\hline
Model &  Precision & Recall & mAP50 & mAP50-95 \\
\hline
DETR&	80.4\%	&62.50\%	&74.8\%	&52.3\% \\
\hline
refined DETR&	85.4\%	&70.2\%	&79.4\%	&58.6\% \\

\hline
\end{tabular}
\end{table}

\section{Conclusion}
This study is based on a comprehensive examination of vehicle detection algorithms under the extreme and variable conditions of Nordic winters. Utilizing the Nordic Vehicle Dataset (NVD), we rigorously evaluated a wide array of state-of-the-art vehicle detection frameworks spanning single-stage, two-stage, and transformer-based architectures. Our analysis revealed the inherent challenges these algorithms face when confronted with the unique environmental factors of the Nordic landscape, particularly in terms of dealing with vehicles fully covered by snow, variable illumination, and other factors.

Through a systematic process of tuning and enhancement tailored to the specific demands of this challenging environment, we demonstrated improvements in the performance of these detection systems. Key strategies such as data augmentation, hyperparameter adjustment, and the innovative use of transfer learning played pivotal roles in this enhancement process. Moreover, in the case of transformer-based frameworks such as DETR, we implemented an approach that initiates region identification using MSER, coupled with a filtering layer grounded in Rough Set theory. These steps demonstrated improvements and refinement in DETR detection performance within challenging circumstances.

This study contributes to the advancement of vehicle detection technologies in challenging weather conditions. It lays the groundwork for future explorations into adaptive, context-aware detection systems capable of maintaining high performance across diverse and dynamic environments.

\printbibliography

\end{document}